\documentclass[conference]{IEEEtran}
\IEEEoverridecommandlockouts

\usepackage[utf8]{inputenc}
\usepackage[T1]{fontenc}
\usepackage[american]{babel}
\usepackage{graphicx}
\usepackage{listings}
\usepackage{float}
\usepackage{amsmath,amssymb,exscale}
\usepackage{algorithm}
\usepackage{algpseudocode}
\usepackage{array}
\usepackage{hyperref}

\begin{document}

\title{Large Language Model Benchmarks in Medical Tasks}

\author{
    \IEEEauthorblockN{
        Lawrence K.Q. Yan\textsuperscript{1}, 
        Qian Niu\textsuperscript{*,2},
        Ming Li\textsuperscript{3},
        Yichao Zhang\textsuperscript{4},
        Caitlyn Heqi Yin\textsuperscript{5},
        Cheng Fei\textsuperscript{6},
        Benji Peng\textsuperscript{7},\\
        Ziqian Bi\textsuperscript{8},
        Pohsun Feng\textsuperscript{9},
        Keyu Chen\textsuperscript{3},
        Tianyang Wang\textsuperscript{10},
        Yunze Wang\textsuperscript{11},
        Silin Chen\textsuperscript{12},\\
        Ming Liu\textsuperscript{13},
        Junyu Liu\textsuperscript{2}
        Xinyuan Song\textsuperscript{14},
        Riyang Bao\textsuperscript{14},
        Zekun Jiang\textsuperscript{15},
        Ziyuan Qin\textsuperscript{14}
    }
    \IEEEauthorblockA{
        \textsuperscript{1}Hong Kong University of Science and Technology
    }
    \IEEEauthorblockA{
        \textsuperscript{2}Kyoto University
    }
    \IEEEauthorblockA{
        \textsuperscript{3}Georgia Institute of Technology
    }
    \IEEEauthorblockA{
        \textsuperscript{4}The University of Texas at Dallas
    }
    \IEEEauthorblockA{
        \textsuperscript{5}University of Wisconsin-Madison
    }
    \IEEEauthorblockA{
        \textsuperscript{6}Cornell University
    }
    \IEEEauthorblockA{
        \textsuperscript{7}AppCubic
    }
    \IEEEauthorblockA{
        \textsuperscript{8}Indiana University
    }
    \IEEEauthorblockA{
        \textsuperscript{9}National Taiwan Normal University
    }
    \IEEEauthorblockA{
        \textsuperscript{10}University of Liverpool
    }
    \IEEEauthorblockA{
        \textsuperscript{11}University of Edinburgh
    }
    \IEEEauthorblockA{
        \textsuperscript{12}Zhejiang University
    }
    \IEEEauthorblockA{
        \textsuperscript{13}Purdue University
    }
    \IEEEauthorblockA{\textsuperscript{14}Emory University}
    \IEEEauthorblockA{\textsuperscript{15}West China Biomedical Big Data Center, West China Hospital, Sichuan University}
    \IEEEauthorblockA{
        *Corresponding Email: niu.qian.f44@kyoto-u.jp
    }
}

\maketitle

\begin{abstract}
With the increasing application of large language models (LLMs) in the medical domain, evaluating these models' performance using benchmark datasets has become crucial. This paper presents a comprehensive survey of various benchmark datasets employed in medical LLM tasks. These datasets span multiple modalities including text, image, and multimodal benchmarks, focusing on different aspects of medical knowledge such as electronic health records (EHRs), doctor-patient dialogues, medical question-answering, and medical image captioning. The survey categorizes the datasets by modality, discussing their significance, data structure, and impact on the development of LLMs for clinical tasks such as diagnosis, report generation, and predictive decision support. Key benchmarks include MIMIC-III, MIMIC-IV, BioASQ, PubMedQA, and CheXpert, which have facilitated advancements in tasks like medical report generation, clinical summarization, and synthetic data generation. The paper summarizes the challenges and opportunities in leveraging these benchmarks for advancing multimodal medical intelligence, emphasizing the need for datasets with a greater degree of language diversity, structured omics data, and innovative approaches to synthesis. This work also provides a foundation for future research in the application of LLMs in medicine, contributing to the evolving field of medical artificial intelligence.
\end{abstract}


\section{Introduction}
Large language models (LLMs) are advanced machine learning systems built on deep learning algorithms, primarily utilizing transformer architectures. By employing attention mechanisms, these models can process extensive amounts of textual data while simultaneously focusing on relevant segments of the input. The remarkable success of ChatGPT by OpenAI has rapidly drawn attention to LLMs, catalyzing a revolution across various industries, including education, customer service, marketing, and notably, healthcare.

In the healthcare sector, the advent of LLMs has led to a surge of innovative applications, ranging from medical education and drug development to clinical trials and disease diagnosis \cite{yang2023large, ghim2023transforming, niu2024text}. These models are now capable of generating reports and facilitating patient-doctor conversations. As LLMs continue to evolve, they can manage increasingly complex data types—not just text and images, but also audio, video, 3D structures, and other modalities. This versatility empowers them to enhance diagnostic accuracy, streamline patient interactions, and support clinical decision-making by integrating diverse information sources.

The growing integration of LLMs in medicine underscores the urgent need for effective evaluation methods to assess their performance across a variety of tasks using structured and diverse datasets. Such datasets are vital for developing LLMs that can tackle complex clinical challenges—like diagnosis and predictive decision support—while simultaneously improving healthcare delivery through enhanced communication between patients and providers.

This manuscript offers a comprehensive overview of benchmark datasets specifically tailored for medical applications of LLMs, which are essential for training and validating these models. The paper categorizes various datasets based on their modalities—text, image, and multimodal—and underscores their significance in areas such as electronic health records (EHRs), doctor-patient interactions, medical question-answering, and medical image analysis. Furthermore, it discusses their applications in various discriminative and generative tasks.

As the healthcare landscape continues to evolve with technological advancements, understanding how to effectively leverage these datasets will be crucial for maximizing the potential benefits of LLMs in enhancing patient care and optimizing clinical workflows. 
\section{Overviews of benchmark datasets}

Benchmark datasets are meticulously curated collections utilized to evaluate and compare the performance of large language models (LLMs). These datasets are typically focused on specific tasks, ensuring they accurately represent the relevant feature space. They are designed to be open, discoverable, and accessible to the research community, facilitating objective comparisons of model performance across various algorithms and techniques.
In the biomedical field, a variety of datasets with different modalities and formats are available for training models across multiple applications. These include text datasets in various formats as well as image-caption datasets featuring medical images from modalities such as X-rays and MRIs. The following sections will classify benchmark datasets sourced from diverse origins according to their data formats and summarize them in tables, highlighting key datasets in the accompanying text.

\subsection{Text benchmark datasets}

Text is the most prevalent type of dataset across all modalities, particularly due to the extensive availability of textual data in the biomedical domain. This abundance is attributed to the ease of collection and annotation, as well as the direct relevance of text in language modeling.
Biomedical textual data encompasses various structured formats, including electronic health records (EHRs), doctor-patient dialogues, open-access literature and abstracts, question-answer pairs, medical guidelines, instructions, and open access literature and abstracts. This wealth of text data serves as a valuable resource for training biomedical large language models (LLMs). Table \ref{tab:text} provides a summary of several popular textual corpora utilized in the training of medical large language models (LLMs).

\subsubsection{\textbf{Electronic Health Records(EHRs)}}\

Electronic health records (EHRs) are digital collections of patient information that provide a comprehensive, real-time view of health data. EHRs include a variety of data types, such as medical history, diagnoses, medications, immunization status, allergies, and laboratory test results. 

These datasets are typically curated by extracting information from local hospitals, de-identifying it in accordance with regulatory standards like HIPAA, and standardizing it into a common schema. The data is then annotated to create labeled datasets for specific tasks.

As electronic systems have become widely adopted, EHRs from different countries and recorded in various languages are increasingly available. MIMIC-III\cite{johnson2016mimic} and its successor MIMIC-IV\cite{johnson2023mimic}  are freely accessible English-language critical care databases containing de-identified health data from over 40,000 patients. This initiative is a collaboration between Beth Israel Deaconess Medical Center (BIDMC) and the Massachusetts Institute of Technology (MIT), featuring clinical notes, discharge summaries, and other structured data. AmsterdamUMCdb\cite{thoral2021sharing} is a Dutch-medium intensive care database endorsed by the European Society of Intensive Care Medicine (ESICM). It contains deidentified health data from over 23,000 ICU and HDU admissions to an academic medical center in Netherlands from 2003-2016. 

The Pediatric Intensive Care (PIC)\cite{zeng2020pic} database is is a Chinese-language resource designed to support research in pediatric critical care; it mirrors the structure of the MIMIC-III database but is tailored for local Chinese data.  The database is sourced from The Children’s Hospital at Zhejiang University School of Medicine with about 13,000 patients and 14,000 ICU stays admitted from 2010–2018. 

Other notable datasets include the German-language HiRID (High-Resolution ICU Dataset)\cite{hyland2020early} that contains data from 34,000 patient admissions to the Department of Intensive Care Medicine at the Bern University Hospital in Switzerland, and the English-language eICU Collaborative Research Database (eICU-CRD)\cite{pollard2018eicu} that includes data from over 200,000 ICU admissions across multiple hospitals in United States.

\subsubsection{\textbf{Doctor-patient dialogues}}\

Doctor-patient dialogues refer to the interactions and exchanges of information between healthcare providers and patients during medical consultations. These dialogues are essential for effective healthcare delivery, as they facilitate understanding, diagnosis, treatment planning, and patient education.

Many available datasets of doctor-patient dialogues are sourced from real conversations collected on online medical consultation platforms. For instance, he iCliniq dataset\cite{li2023chatdoctor} consists of 10,000 authentic conversations between patients and doctors from the iCliniq online consultation platform. A similar dataset, HealthCareMagic-100k\cite{li2023chatdoctor}, consists of 100,000 real conversations from the HealthCareMagic platform. Additionally, datasets are available in languages other than English. The Meddialog dataset \cite{he2020meddialog} contains 1.1 million Chinese consultations sourced from haodf.com, covering nearly all medical specialties and including services like medical consultations and appointment scheduling. The IMCS-21\cite{chen2023benchmark} dataset is another popular Chinese-medium doctor-patient dialogue dataset contains over 60,000 medical dialogue sessions with data collected from from Muzhi (http://muzhi.baidu.com), a Chinese online health community offering professional medical consulting service for patients. As these datasets sourced data from real online consultation platform, they require deidentification of both patients and doctors to protect privacy and have strict data regulation according to the HIPAA. 

Other datasets are created using large language models (LLMs) like ChatGPT to generate synthetic dialogues based on medical documents such as clinical notes. This approach allows for the creation of realistic dialogues without compromising patient confidentiality and facilitates the generation of multi-turn dialogue datasets crucial for modeling real-world healthcare conversations. The NoteChat dataset\cite{wang2023notechat} isan example of a synthetic dialogue dataset conditioned on 167k case reports created using a cooperative multi-agent framework that leverages LLMs to generate patient-physician dialogues based on 167,000 case reports from the PubMed Central repository. \cite{roberts2001pubmed}. SynDial\cite{das2024synthetic} is another synthetic patient-physician dialogues dataset generated through iterative refinement of LLM outputs using electronic health records from the MIMIC-IV dataset.

\subsubsection{\textbf{Medical question answering}}\

Medical question answering datasets consist of collections of questions and answers pertinent to the medical and healthcare fields. These datasets are curated to evaluate the capabilities of language models and question-answering (QA) systems in understanding and responding to a wide range of medical inquiries, thereby facilitating the research and development of advanced medical QA technologies.

These datasets vary in format, typically encompassing semantic, yes/no, multiple-choice, and open-ended question answering. Data is often sourced from literature, clinical exam questions, online health forums, or synthesized from existing medical texts or abstracts found in research papers and textbooks using natural language processing techniques.

\paragraph{Semantic Question Answering (SQA)}\

Semantic question answering (SQA) focuses on understanding the meaning of questions and their context to provide accurate answers, rather than merely matching keywords. SQA datasets usually datasets contain a wider variety of question types and include ground-truth answers that are verified against a knowledge base. This verification allows for precise measurement of SQA system performance by comparing outputs to these established correct answers.

A popular SQA datasets in biomedical applications is the BioASQ-QA\cite{krithara2023bioasq} dataset, which is a manually curated question answering corpus derived from the BioASQ challenges that have been focused on biomedical semantic indexing and question answering since 2013. The dataset contains factoid, list, yes/no, and summary questions in English, along with golden standard exact answers and ideal answers in effect summaries which are referenced to documents indexed for MEDLINE.

\begin{table*}[htbp]
\centering
\caption{Overview of textual benchmark datasets in medical AI}
\begin{tabular}{p{0.3\textwidth}  p{0.2\textwidth}  p{0.2\textwidth} p{0.2\textwidth}}
\hline
\textbf{Type} & \textbf{Benchmarks (Year)} \\
\hline

\textbf{EHR} 
&AmsterdamUMCdb (2021) \cite{thoral2021sharing}
&eICU-CRD (2018)
\cite{pollard2018eicu}
&HiRID (2020)
\cite{hyland2020early}\\ 

&MD-HER (2023) \cite{wang2023clinicalgpt} 
&MIMIC-III (2016) \cite{johnson2016mimic}
&MIMIC-IV (2023)\cite{johnson2023mimic}\\
&MIMIC-ED (2023) \cite{https://doi.org/10.13026/5ntk-km72}
&PIC (2020) \cite{zeng2020pic}
&PMC-Patients (2022)\cite{zhao2022pmc}\\
\\

\textbf{Medical conversations} 
&BianQueCorpus(2023)\cite{chen2023bianque} 
&CMtMedQA (2024)\cite{yang2024zhongjing} 
&GenMedGPT-5k (2023)\cite{li2023chatdoctor}\\
&HealthCareMagic-100k (2023)\cite{li2023chatdoctor}
&Icliniq-10K (2023)\cite{li2023chatdoctor}
&IMCS-21/IMCS-V2 (2023)\cite{chen2023benchmark} \\
&MedDG (2022)\cite{liu2022meddg} 
&Meddialog (2020)\cite{he2020meddialog}
&Notechat (2023)\cite{wang2023notechat}\\
&Psych8K (2023)\cite{liu2023chatcounselor}
\\ \\

\textbf{Medical question answering(QA)} 
&\textbf{Semantic QA:} \\
&emrQA (2018)\cite{pampari2018emrqa}
&TREC (2001)\cite{hovy2001toward}
&Medication\_QA\_MedInfo2019 (2019)\cite{abacha2019bridging}\\
&cMedQA (2017)\cite{zhang2017chinese}
&Huatuo-26M (2023)\cite{li2023huatuo}
&MedQuAD (2019)\cite{ben2019question} \\
&BioASQ-QA (2023)\cite{krithara2023bioasq}
&SQuAQ (2016)\cite{rajpurkar2016squad}
\\
\\

&\textbf{MCQA:}\\
&CEval (2024)\cite{huang2024c}
&CMExam (2024)\cite{liu2024benchmarking}
&MultiMedQA (2022)\cite{singhal2022large}\\
&PubMedQA (2019)\cite{jin2019pubmedqa}
&USMLE (since 1994)\cite{usmleHomeUnited}
&XMedBench (2024)\cite{wang2024apollo}\\ \\

&\textbf{OpenQA:}\\
&MedQA (2020)\cite{jin2021disease}\\ \\

&\textbf{Multiple formats:}\\
&MedMCQA (2022)\cite{pal2022medmcqa} &BiMed1.3M (2024)\cite{pieri2024bimedix}\\
\\ \\

\textbf{Literatures, abstracts and derivatives} 
&\multicolumn{3}{l}{\textbf{Academic literatures:} }\\
&BC4GO (2014)\cite{van2014bc4go}
&BC5CDR (2016)\cite{li2016biocreative}
&BC4CHEMDNER (2015)\cite{krallinger2015chemdner}\\
&ChemProt (2018)\cite{krallinger2017overview}
&GENIA (2003)\cite{kim2003genia}
&HoC (2016)\cite{baker2016automatic}\\
&JNLPBA (2004)\cite{collier2004introduction}
&NCBI-disease (since 2003)\cite{dougan2014ncbi}
&PubMed Central (PMC) (since 2000)\cite{roberts2001pubmed}\\
&S2ORC (2019)\cite{lo2019s2orc}
TREC-COVID (2020)\cite{voorhees2021trec}
\\ \\
&\multicolumn{3}{l}{\textbf{Abstracts of patents and clinical trails:} }\\
&CheF (2024)\cite{kosonocky2024mining}
&CHEMDNER-patents (2016)\cite{zhang2016chemical}
&EBM-NLP (2018)\cite{nye2018corpus}\\
&USPTO (since 2015)\cite{huang2004international}\\
\\

\textbf{NLP tasks}
&ADE (2012)\cite{gurulingappa2012development}
&AutoMeTS (2020)\cite{van-etal-2020-automets}
&BioNLI (2022)\cite{bastan2022bionli}\\
&BioRED (2022)\cite{luo2022biored}
&BIOSSES (2017)\cite{souganciouglu2017biosses}
&Cadec (2015)\cite{karimi2015cadec}\\
&CliCR (2018)\cite{vsuster2018clicr}
&CMeEE (2021)\cite{zhang2021cblue}
&DDI (2013)\cite{herrero2013ddi}\\
&EU-ADR (2012)\cite{van2012eu}
&GAD (2004)\cite{becker2004genetic}
&GDA (2019)\cite{wu2019renet}\\
&HunFlair2 (2024)\cite{weber2021hunflair}
&i2b2 (since 2004)\cite{uzuner20112010}
&Medical Abstracts (2022)\cite{schopf2022evaluating}\\
&MEDIQA NLI (2019)\cite{abacha2019overview}
&MEDIQA QA (2019)\cite{abacha2019overview}
&MEDIQA RQE (2019)\cite{abacha2019overview}\\
&MedNLI (2019)\cite{romanov2018lessons}
&MedQSum (2023)\cite{10373720}
&MedSTS (2018)\cite{wang2020medsts}\\
&MentSum (2022)\cite{sotudeh-etal-2022-mentsum}
&MeQSum (2019)\cite{abacha2019summarization}
&MultiCochrane (2023)\cite{joseph2023multilingual}\\
&n2c2 (since 2019)\cite{stubbs2019cohort}
&NFCorpus (2016)\cite{boteva2016full}
&NLM-Chem-BC7 (2022)\cite{islamaj2022nlm}\\
&OHSUMED (1994)\cite{hersh1994ohsumed}
&PGR (2019)\cite{sousa2019silver}
&WNUT-2020 Task 2 (2020)\cite{nguyen2020wnut}\\
\\

\textbf{Large composite corpus}
&ApolloCorpora (2024)\cite{wang2024apollo}
&BEIR (2021)\cite{thakur2021beir}
&BigBIO (2022)\cite{fries2022bigbio}\\
&BLUE (2019)\cite{peng2019transfer}
&BLURB (2021)\cite{gu2021domain}
&BoX (2022)\cite{parmar2022boxbart}\\
&CBLUE (2022)\cite{zhang2021cblue}
&ChiMed-CPT (2024)\cite{tian2023chimed}
&MedBench (2024)\cite{liu2024medbench}\\
&Medical Meadow(2023)\cite{han2023medalpaca}
&PMC-LLaMA (2023)\cite{wu2023pmc}
&PromptCBLUE (2023)\cite{zhu2023promptcblue}\\
&The pile (2020)\cite{gao2020pile}\\
\\

\hline
\end{tabular}
\label{tab:text}
\end{table*} 
\paragraph{Multi-Choice Question Answering (MCQA)}\

MCQA datasets present the questions in a multiple-choice format, often with four or five answer options.  This structure not only tests factual recall but also evaluates the ability to apply knowledge in various contexts, assessing critical thinking and clinical reasoning skills.

Due to the prevalence of MCQA in examination settings, many datasets are derived from established medical exams. The MedQA dataset\cite{jin2021disease} sources questions from the United States Medical Licensing Examination (USMLE), covering general medical knowledge. It includes 11,450 questions in the development set and 1,273 questions in the test set,  each with four or five answer choices. Another example is the MedMCQA\cite{pal2022medmcqa} which draws from the Indian medical entrance exams (AIIMS/NEET), and covers 2,400  healthcare topics across 21 medical subjects. It has over 187,000 questions in the development set and 6,100 questions in the test set, with 4 answer choices per question.

A less common variant of MCQA can be found in the PubMedQA dataset, which adopt a yes/no/maybe format for answers. This biomedical QA dataset is derived from PubMed abstracts, where each question in the dataset is paired with an abstract from a PubMed article excluding its conclusion, which serves as the long answer. To answer the questions in the PubMedQA\cite{jin2019pubmedqa} dataset, models must effectively parse and reason through the abstract to arrive at the correct answer, highlighting the importance of understanding context rather than just retrieving facts. The binary classification approach also emphasizes the model's ability to discern conclusions based on abstract information rather than simply selecting from multiple options. 

\paragraph{Open Question Answering (OpenQA)}\

Datasets for Open-domain Question Answering (OpenQA) usually do not include text explicitly labeled as answers; instead, they are associated with a large knowledge base, such as Wikipedia or other unstructured text sources. Models designed for OpenQA generally follow a retrieve-then-read paradigm, where they first retrieve relevant documents or data from a vast corpus and then perform machine comprehension to generate answers based on the retrieved information. These datasets often adopt a Question-Answer-Context triplet format, which aids the model in understanding the relationships between the different components.

As a result, some of the aforementioned datasets are also utilized for training models in the OpenQA task. For example, the MedQA\cite{jin2021disease} dataset has a document collection constructed from 18 english textbooks and 33 simplified Chinese textbook, the BioASQ-QA dataset\cite{krithara2023bioasq} has materials indexed for MEDLINE; and PubMedQA has relevant paper abstract, associated with the question and answer pairs.

\subsubsection{\textbf{Literatures, abstracts and derivatives}}\

Biomedical literature provides a vast repository of specialized knowledge, covering diverse topics such as diseases, treatments, drug interactions, and clinical trials. With their key findings and methodologies summarized in abstracts, both the literature and the abstracts are valuable resources for training large language models (LLMs) in the biomedical domain.

These datasets enable LLMs to learn the specific language, terminology, and context relevant to the biomedical field, enhancing their ability to generate accurate and contextually appropriate outputs.They also improve the models' capacity to identify entities such as genes, proteins, diseases, and medications within texts and to understand the complex relationships among them. Training LLMs with both the literature and their corresponding abstract also allows LLMs to develop skills in summarization and information extraction and improve the performance of models in tasks such as information retrieval and question-answering.

PubMed and its subset PubMed Central (PMC)\cite{roberts2001pubmed} are both popular sources of literature and abstracts. While PubMed is a comprehensive database that includes references and abstracts from various biomedical literature, including more than 37 million articles from journals that may not be freely accessible; PMC is a free digital repository that specifically archives over 5 million full-text articles in the biomedical and life sciences. 

S2ORC\cite{lo2019s2orc} is another popular dataset consisting of open access academic papers from the Semantic Scholar literature corpus with domain not only on the biomedical field. The dataset consists of rich metadata, paper abstracts, resolved bibliographic references, as well as structured full text for 8.1M open-access papers.

In addition to academic literature, documents such as patents and clinical trial reports play a crucial role in the biomedical innovation process, particularly when healthcare products are poised for commercial launch. The capabilities of large language models in text summarization can significantly accelerate the preparation of these materials. Also, such documents often reveal important relationships between medical drugs or treatments and their outcomes, making the availability of datasets in these formats essential.

The Chemical Function (CheF)\cite{kosonocky2024mining}  and USPTO Patent dataset\cite{huang2004international} are two important datasets in this category. The CheF dataset contains 631,000 molecule-function pairs derived from approximately 100,000 unique molecules across 188,000 patents, created by using LLMs to summarize patent titles, abstracts, and descriptions into functional labels. The USPTO Patent Dataset includes over 8 million patents, structured into various fields such as title, abstract, claims, and descriptions. These datasets serves as a foundational resource for developing models that can generate and retrieve patent information effectively, facilitating research in biomedicine. 

MedReview\cite{zhang2024closing}  dataset consists of 8161 pairs of meta-analysis results from randomized controlled trails (RCTs) along with narrative summaries from the Cochrane Library, published on 37 topics between April 1996 and June 2023. This data set can be used to train LLMs to systematically review randomized controlled trials (RCTs) and summarize medical evidence, which is crucial in making healthcare decisions. 

\subsubsection{\textbf{Datasets curated for a specific NLP task}}\

Apart from textual datasets curated as a collection of documents in a specific format, datasets curated for specific natural language processing (NLP) tasks are also available. These datasets are curated to enhance the performance of models in specialized areas, such as named entity recognition (NER), relation extraction, and clinical text generation. Tailoring datasets for specific tasks can significantly improve model performance metrics such as F1 scores, precision, and recall. 

Datasets originated from the MEDIQA\cite{abacha2019overview} initiative are great examples. The MEDIQA refers to a series of shared tasks and challenges focused on evaluating the capabilities of large language models (LLMs) in the biomedical domain, particularly in clinical contexts.The MEDIQA NLI, MEDIQA QA and MEDIQA RQE are datasets sourced from the initiative  that focuses on Natural Language Inference (NLI), Question Answering (QA) and Recognizing Question Entailment (RQE), repectively.

\subsubsection{\textbf{Large composite corpus}}\

Instead of training LLMs on a single dataset for a specific task, researchers usually combine multiple datasets, with data from different sources, modalities and formats, into a larger corpus for the training of the models. By integrating multiple datasets from diverse sources, researchers can enhance model performance through Multi-Task Learning (MTL), which allows for simultaneous training on different related tasks. Notable examples of large composite biomedical benchmark datasets include the Biomedical Language Understanding Evaluation (BLUE)\cite{peng2019transfer}, BLURB (Biomedical Language Understanding and Reasoning Benchmark)\cite{gu2021domain}, BoX\cite{parmar2022boxbart}, The pile\cite{gao2020pile} and BigBIO\cite{fries2022bigbio} datasets, which comprise increasing numbers of datasets, task categories, and languages. 

\subsubsection{\textbf{The evolution of textual benchmarks}}\

Early advancements in large language models (LLMs), particularly those based on the GPT and BERT architectures, primarily utilized general-purpose datasets. While these models demonstrated impressive capabilities in generating and understanding human language, they lacked the specificity required for medical applications. As researchers began to explore the potential of LLMs in healthcare, it became clear that general datasets could not adequately address the complexities inherent in medical language and context.

The need arose for models capable of understanding clinical terminology, interpreting unstructured clinical notes, and assisting in clinical decision-making. This realization led to the development of medical-specific datasets. Electronic health records (EHRs) emerged as a critical resource, offering vast amounts of unstructured clinical data that could be leveraged to train LLMs tailored for healthcare applications. For instance, datasets like NYU Notes, which encompass millions of clinical notes, exemplify this shift by providing rich, domain-specific content for model training.

The introduction of clinician-generated datasets, such as MedAlign, marked a significant advancement in this area. These datasets comprise natural language instructions curated by healthcare professionals to enhance the relevance and applicability of LLMs in real-world clinical settings.
As LLM capabilities evolved to handle modalities beyond text and embrace multimodalities, the integration of multi-modal datasets became increasingly important. This trend is illustrated by datasets like MedPix 2.0, which links clinical reports with corresponding imaging data, thereby enhancing the ability of LLMs to perform complex analyses pertinent to patient care. Additionally, Quilt combines image-text pairs to improve LLM training by utilizing diverse data sources.

In summary, the shift from general NLP datasets to medical-specific datasets signifies an increasing recognition of the unique demands of healthcare applications, paving the way for more effective and relevant large language models (LLMs) in clinical settings. The following section will discuss benchmark datasets that consist of both images and accompanying caption text.

\subsection{Captioned imagery benchmarks}
The development of large language models (LLMs) has significantly advanced traditional methods for processing text benchmark datasets. Recent improvements in multimodal LLMs have expanded their capabilities, allowing them to effectively manage complex datasets, especially in the medical field. These advancements are crucial for integrating various data types—such as clinical text and medical images—which are essential for thorough medical analysis and decision-making. Imagery datasets are particularly important, as medical diagnostics often rely on both visual and textual information. By using captioned imagery datasets, LLMs can combine visual data with textual descriptions, enhancing their understanding of complex clinical scenarios and improving diagnostic accuracy. 

In recent years, captioned imagery benchmarks have emerged as critical tools for evaluating the performance of LLMs in multimodal tasks, especially in vision-language understanding. These benchmarks assess models' abilities to generate coherent and contextually appropriate descriptions of images, which is foundational for various downstream applications such as automated image annotation, accessibility tools for visually impaired users, and improved human-computer interaction.

In medical applications, Medical Image Captioning (MIC) combines the power of multimodal LLMs with specialized medical knowledge to produce accurate descriptions of medical images. This task is challenging due to the complexity of medical imagery; models must understand intricate anatomical structures, identify pathological conditions, and communicate findings using professional terminology. Table \ref{tab:img} summarizes several popular imagery corpora used in training medical LLMs.

\subsubsection{\textbf{Examples of Medical Imagery Benchmarks}}\
Several notable benchmarks and datasets have been developed to advance research in medical image captioning:

1. IU X-Ray Dataset: This dataset, derived from the Indiana University Chest X-Ray Collection, contains over 7,000 chest X-ray images paired with radiological reports. It serves as a primary benchmark for developing and evaluating models capable of generating textual descriptions for chest radiographs.

2. MIMIC-CXR Dataset: Part of the larger MIMIC (Medical Information Mart for Intensive Care) database, MIMIC-CXR provides over 377,000 chest X-rays associated with 227,835 imaging studies. This large-scale dataset is crucial for training robust models that can handle diverse pathological conditions and varying image qualities.

3. Radiology Objects in COntext (ROCO) Dataset: This dataset consists of over 81,000 radiology images from various modalities (e.g., CT, MRI, X-ray) paired with captions extracted from scientific papers. ROCO is particularly useful for training models to generate captions that align with scientific literature standards.

4. VQA-RAD Dataset: While primarily designed for visual question answering, this dataset of 315 medical images and 3,515 question-answer pairs can also be adapted for image captioning tasks, especially for generating focused descriptions based on specific aspects of medical images.

5. CheXpert Dataset: Containing 224,316 chest radiographs of 65,240 patients, CheXpert is a large public dataset for chest radiograph interpretation. While not specifically designed for captioning, it can be leveraged to enhance the performance of medical image captioning models, especially in identifying and describing multiple coexisting conditions.

\begin{table*}[htbp]
\centering
\caption{Overview of captioned imagery benchmark datasets in medical AI}
\begin{tabular}{p{0.2\textwidth}  p{0.23\textwidth}  p{0.23\textwidth} p{0.23\textwidth}}
\hline
\textbf{Type} & \textbf{Benchmarks (Year)} \\
\hline

\textbf{2D image} 
&\textbf{CT:}\\
&COVID19-CT-DB (2021)\cite{shakouri2021covid19}
&CT-Kidney (2022)\cite{islam2022vision}
&DeepLesion (2017)\cite{yan2017deeplesion}\\
&LIDC-IDRI (2015)\cite{https://doi.org/10.7937/k9/tcia.2015.lo9ql9sx}\\
\\

&\textbf{Clinical image:}\\ 
&PAD-UFES-20 (2020)\cite{pacheco2020pad}\\ \\

&\textbf{Ultrasound:}\\ 
&BUSI (2019)\cite{al2020dataset}\\ \\

&\textbf{X-ray:}\\ 
&ChestX-Det10 (2020)\cite{liu2020chestx}
&ChestX-ray14 (2017)\cite{wang2017chestx}  
&CheXpert (2019) \cite{irvin2019chexpert}\\
&CMMD2022 (2019) \cite{cai2019breast}
&COVID-19-Radio (2020)\cite{chowdhury2020can} 
&Covid-CXR2 (2022) \cite{pavlova2022covid}\\
&CXR8 (2017) \cite{wang2017chestx}
&Montgomery County X-ray (2021)\cite{jaeger2014two} 
&MURA (2017) \cite{rajpurkar2017mura}\\
&NIH Chest X-ray (2017)\cite{wang2017chestx}
&PadChest (2020) \cite{bustos2020padchest}
&RSNA(2017)\cite{shih2019augmenting}\\
&SIIM-ACR(2019) \cite{filice2020crowdsourcing}
&Vertebrase-Xray (2022) \cite{fraiwan2022using}
&VinDr-BodyPartXR (2021)\cite{pham2021dicom}\\
&VinDr-CXR (2020)\cite{nguyen2022vindr}
&VinDr-Mammo (2022)\cite{nguyen2023vindr}
&VinDr-Multiphase (2022)\cite{dao2022phase}\\ 
&VinDr-PCXR (2022) 
&VinDr-RibCXR (2021)\cite{nguyen2021vindr}
&VinDr-SpineXR (2021)\cite{nguyen2021vindr}
\\ \\

&\textbf{MRI:}\\ 
&Brain-Tumor (2021) 
&Brain-Tumor-17 (2022) 
&CE-MRI (2015)\cite{cheng2015enhanced}\\
&MRNet (2018)\cite{bien2018deep}
\\ \\

&\textbf{Multimodality:}\\ 
&RadImageNet (2022)\cite{mei2022radimagenet}
\\ \\

\textbf{2D image + text} &
\textbf{CT:}\\ 
&RadGenome-ChestCT (2024)\cite{zhang2024radgenome}
\\ \\

&\textbf{Histopathology:}\\ 
&QUILT-1M (2023)\cite{ikezogwo2024quilt}
\\ \\

&\textbf{X-ray:}\\ 
&CBIS-DDSM (2017)\cite{lee2017curated}
&IU X-ray (2019)\cite{demner2016preparing}
&MIMIC-CXR (2020)\cite{johnson2019mimic}
\\
&OpenI Chest X-rays (2012)\cite{nihOpeniNgWeb}
\\

&
\textbf{Multimodality:}\\ 
&PEIR GROSS (2019)\cite{pavlopoulos2019survey}
&ROCO (2018) \cite{pelka2018radiology}
&MedPix (since 2016)\\
&ImageCLEF (since 2003) \cite{de2016overview}
&PMC-Fine-Grained-46M (2023) \cite{zhang2023biomedclip}
&PMC-15M (2023)\cite{zhang2023biomedclip}\\ \\

&\textbf{Visual QA datasets:}\\ 
&OVQA (2022) \cite{huang2022ovqa}
&Path-VQA (2020)\cite{he2020pathvqa}
&PMC-CaseReport (2023)\cite{zhang2023pmc}\\
&PMC-VQA (2024)\cite{zhang2023pmc}
&RadVisDial (2019)\cite{kovaleva2019visual}
&VQA-Med(since 2018)\cite{ben2019vqa,ben2021overview}\\
&VQA-RAD (2018)\cite{lau2018dataset} 
&Slake-VQA (2021)\cite{liu2021slake} 
&Kvasir-VQA (2024)\cite{gautam2024kvasir}\\
\\

\textbf{3D image} &
\textbf{MRI:}\\ &
ISPY1(2016)\cite{https://doi.org/10.7937/k9/tcia.2016.hdhpgjlk}
&ISPY2(2022)\cite{https://doi.org/10.7937/tcia.d8z0-9t85}
&OASIS-3 (2019)\cite{lamontagne2019oasis}\\
&BraTS(2021)\cite{baid2021rsna}\\
\\ \\

&\textbf{CT:}\\ 
&CC-CCII (2020)\cite{zhang2020clinically}
&CT-GAN (2019) \cite{mirsky2019ct}
&CT-RATE (2024)\cite{hamamci2024foundation}\\
&ICH2020 (2020)\cite{hssayeni2020intracranial}
&LIDC-IDR (2011)\cite{armato2011lung} 
&LNDb (since 2019)\cite{pedrosa2019lndb}\\
&MosMedData (2020)\cite{morozov2020mosmeddata}
&RAD-ChestCT (2020)\cite{draelos2021machine}
\\ \\

&\textbf{Multimodality:}\\ 
&NSCLC-Radiomics (2018) \cite{bakr2018radiogenomic}
&RP3D-DiagDS(2023)\cite{zheng2023large}\\
\\

\textbf{2+3D image}&
\textbf{X-ray:}\\ 
&KneeMRI(2017)\cite{vstajduhar2017semi}\\
\\

&\textbf{Multimodality:}\\ 
&ADNI1 (2004)
&MedMD (2023)\cite{wu2023towards}
&MedMNIST (2021)\cite{yang2023medmnist}\\ 
&MedTrinity-25M (2024)\cite{xie2024medtrinity}
&MultiMedBench (2023)\cite{tu2024towards}
&TCGA (2016)\\ 
&TCGA-BRCA (2016)\cite{lingle2016cancer}
&TCGA-LUAD (2016)\cite{https://doi.org/10.7937/k9/tcia.2016.jgnihep5}
&TCGA-LUSC (2020)\cite{https://doi.org/10.7937/k9/tcia.2016.tygkkfmq}\\
\\ \\

\hline
\end{tabular}
\label{tab:img}
\end{table*} 
\subsubsection{\textbf{Domain-Specific Image-Caption Benchmarks}}\
As image-captioning expands, the need for domain-specific benchmarks is growing. In medical applications, large multimodal datasets are frequently used to train models that generate captions from complex medical images like X-rays, CT scans, and MRIs. These captions play a vital role in diagnostics, providing detailed and accurate descriptions that help healthcare professionals make informed decisions. One notable benchmark is the MedICaT dataset, which evaluates models' abilities to describe medical images by combining textual reports with visual data \cite{he2023towards}.

\subsubsection{\textbf{Image-Caption Benchmarks in Multimodal Tasks}}\
With the emergence of multimodal models like GPT-4V, BLIP, and CLIP, there is an increasing demand for benchmark datasets to evaluate how effectively these models can process complex multimodal data. This includes context-aware captions and reasoning across various domains.

For example, the Mementos benchmark \cite{wang2024mementos}  focuses on image sequences instead of single static images, enabling the assessment of models' abilities in temporal reasoning—tracking changes and interpreting behaviors throughout a series of images. Similarly, the CODIS benchmark evaluates how well models can maintain context across multiple images, which is essential for tasks that require a coherent understanding of visual scenes over time.\cite{he2023towards, wang2024mementos}.

This shift towards dynamic image-captioning is particularly relevant in real-world applications, such as medical imaging. Techniques like Positron Emission Tomography (PET) and Single Photon Emission Computed Tomography (SPECT) capture sequential images over time, highlighting the need for models to understand and describe non-static visual inputs.

Dynamic understanding and captioning are crucial in fields like medical imaging and robotics. In these domains, models must account for changing visual contexts and generate descriptions that accurately reflect ongoing events, interactions, and outcomes. For example, in robotics, a model may need to detail not only the objects present but also how a robotic arm interacts with them over time, making temporal comprehension vital for effective task execution.\cite{wang2024mementos}.

\subsubsection{\textbf{The Evolution of Image-Caption Benchmarks}}\
Traditionally, image-caption benchmarks have focused on static datasets consisting of single images paired with human-annotated captions. Early benchmarks like MSCOCO and Pascal VOC aimed to generate descriptive captions that captured key objects, actions, and relationships within a single image. The success of these benchmarks spurred the development of more sophisticated models capable of understanding both visual and textual inputs.

With advancements in large language models (LLMs) and multimodal LLMs (MLLMs), the ability to process and generate language from visual inputs has expanded beyond isolated tasks. These models are now evaluated not only on their grammatical accuracy but also on their capability to understand complex scenes, identify relationships between objects, and perform high-level reasoning about images. This evolution necessitates new benchmarks that address a broader range of visual reasoning challenges, including temporal dynamics, object behaviors, and domain-specific imagery.

In the medical field, image captioning has become increasingly important. Medical Image Captioning (MIC) combines LLMs with specialized medical knowledge to produce accurate descriptions of medical images. This task presents unique challenges, such as the need for precise medical terminology and the ability to convey uncertainty in diagnoses. Effective evaluation metrics for MIC extend beyond traditional measures to include clinical accuracy, completeness, and consistency with established medical reporting standards.

As research progresses, benchmarks for captioned medical image are expected to evolve by incorporating diverse imaging modalities and rare conditions. This advancement aims to enhance the capabilities of MLLMs in healthcare applications, ultimately improving diagnostic processes and supporting clinical decision-making through automated and accurate interpretations of medical images.
 
\subsection{Medical benchmarks in other modalities}\
Multimodal large language models (MLLMs) can simultaneously process various data types, enabling a more comprehensive understanding of complex medical scenarios. For example, they can analyze clinical notes, diagnostic images, and patient audio, leading to more accurate diagnoses compared to unimodal systems that rely solely on text or images.

To effectively train an MLLM, diverse datasets beyond just text and images are essential. These include audio, video, ECG data, and various omics datasets\cite{alsaad2024multimodal}. Such multi-modal datasets enhance the model's ability to understand intricate medical scenarios and improve predictive accuracy. The integration of these different modalities fosters richer insights and better contextual understanding in healthcare applications. Table \ref{tab:OthMod} summarizes several other-modality benchmark datasets used in training medical MLLMs.

\subsubsection{\textbf{Video}}\

Videos are sequences of images (frames) that show motion and changes over time. These temporal information can be challenging to work with because it is hard to caption and align, and models must effectively understand the dynamics of movement. Additionally, videos may have redundant visual information, making it difficult to extract important features. In medical applications, video datasets can be generally classified into 4 categories, including surgical ,patient interaction, video-based medical imaging and Human Posture datasets.

\paragraph{Surgical videos}\

These datasets capture various surgical procedures, annotated with key steps and outcomes. These can be used to train models for surgical training simulations or to analyze techniques\cite{waisberg2024openai}. For example, OphNet\cite{hu2024ophnet} is a large-scale video benchmark specifically designed for ophthalmic surgical workflow understanding. It contains more than 2000 videos across 66 types of surgeries, with detailed annotations covering 102 surgical phases and 150 operations. Similary, Cholec80\cite{twinanda2016endonet} contains 80 laparoscopic cholecystectomy surgery videos performed by 13 surgeons. These datasets are valuable resource for studying surgical workflows and techniques and improving action understanding and model validation.

\paragraph{Patient interaction videos}\

These datasets contain videos documenting patient consultations or examinations can help train MLLMs to interpret both verbal and non-verbal cues in a clinical setting. A notable example is the Bristol Archive Project dataset\cite{jepson2017one}. This dataset features 327 video-recorded primary care consultations collected in and around the Bristol between 2014 - 2015, along with coded transcripts.

\paragraph{Video-based medical imaging}\

These datasets include video sequences of imaging techniques, such as ultrasound or endoscopy, annotated with diagnostic information. Notable examples include the C3VD and the S.U.N. Colonoscopy datasets. The C3VD dataset\cite{bobrow2023} comprises 22 registered videos with paired ground truth depth, surface normals, optical flow, occlusion, six degree-of-freedom pose, coverage maps, and 3D models. This allows  comprehensive analysis and training of algorithms in 3D reconstruction and depth estimation tasks. The S.U.N. Colonoscopy Datasets\cite{misawa2021development} includes 49,136 annotated frames from 100 different polyps, providing detailed visual data critical for research in automated polyp detection and classification.The S.U.N. datasets serve as a benchmark for developing and evaluating machine learning models aimed at improving polyp detection rates during colonoscopies. By providing high-quality annotated data, the dataset supports the training of deep learning models to enhance diagnostic accuracy and reduce missed detections in clinical settings.

\paragraph{Human Posture}\

Human pose estimation is essential for intelligent systems that require an understanding of human activities\cite{chen2020monocular, peng2024emerging}. Several benchmarks have been developed to advance this field, each addressing unique challenges. The MVOR dataset\cite{srivastav2018mvor} is the first to offer multi-view RGB-D data from real clinical environments, captured during surgical procedures. It includes synchronized frames from multiple cameras, with annotations for 2D/3D poses and human bounding boxes. Challenges such as occlusions and privacy blurring are present, impacting baseline performance, thus encouraging the development of robust models suited for complex, real-world scenarios. The MPII Human Pose dataset\cite{andriluka20142d} is known for its diversity, covering over 800 human activities. It provides extensive labels, including joint positions, 3D orientations, and occlusion information. The inclusion of adjacent video frames allows for the use of temporal information, enhancing pose estimation. This benchmark sets a standard for evaluating models across a wide variety of real-world activities. The PoseTrack benchmark\cite{andriluka2018posetrack} focuses on video-based multi-person pose estimation and tracking. It introduces tasks for single-frame and video-based pose estimation, as well as articulated tracking, offering a large dataset with labeled person tracks. A centralized evaluation platform enables objective comparison of methods, driving progress in consistent tracking across video sequences. These benchmarks collectively address the complexities of human posture/pose estimation, from handling occlusions to maintaining temporal consistency, fostering more robust and context-aware solutions.

\subsubsection{\textbf{Audio}}\

There are primarily two types of audio datasets used in medical large language models (MLLMs). The first type consists of recordings of medical conversations between doctors and patients, while the second type includes audio recordings for medical classifications, such as breath sounds\cite{hsu2021benchmarking} and heart sounds\cite{abbas2024artificial}.
The former provides rich contextual data that enhances the model's understanding of medical terminology and patient interactions, improving the adaptability of speech recognition systems across various accents and dialects, which is crucial in diverse healthcare settings. In contrast, the latter offers alternative inputs for disease classification, thereby enhancing diagnostic accuracy.

\paragraph{Audio medical conversation}\

While there are numerous commercially available audio datasets for medical conversations, only a few open-access benchmark datasets exist. One notable example is the synthetic patient-physician interviews created by Fareez et al. (2022), structured in the format of Objective Structured Clinical Examinations (OSCE) and specifically targeting respiratory cases. \cite{fareez2022dataset} This dataset contains 272 audio files in MP3 format, along with their transcriptions categorized by disease type. Another example is the Kaggle dataset on medical speech\cite{kaggleMedicalSpeech}, which includes 8.5 hours of audio utterances paired with text for common medical symptoms.

\paragraph{Medical classification}\

Audio recordings can also be instrumental in symptom recognition, particularly for breath and heart sounds.  For instance,  the HF\_Lung\_V1 database \cite{hsu2021benchmarking} is curated for detecting breath phases and adventitious sounds, which  consists of 9,765 audio files of lung sounds, each lasting 15 seconds. It features detailed labeling for breath phases (inhalation and exhalation) and various adventitious sounds, such as wheezes, stridor, rhonchi, and crackles. Another relevant dataset comes from the PASCAL Heart Sounds Challenge\cite{abbas2024artificial}, which provides several hundred real heart sounds collected from an iPhone app and a digital stethoscope in noisy environments. These sounds are classified into specific categories, enhancing models' ability to detect cardiac pathologies.

\subsubsection{\textbf{Electrocardiogram(ECG)}}\

An electrocardiogram (ECG or EKG) is a diagnostic test that measures the electrical activity of the heart over time, capturing voltage changes associated with the heart's rhythmic contractions. This quick, painless, and noninvasive procedure involves placing electrodes on the chest, arms, and legs to record the heart's electrical signals. ECGs are essential for diagnosing various cardiac conditions, including myocardial infarction, arrhythmias, and unstable angina pectoris. Standard ECG leads are labeled as I, II, III, aVF, aVR, aVL, V1, V2, V3, V4, V5, and V6 and are routinely obtained upon patient admission to emergency departments or hospital floors. They may be repeated for patients exhibiting cardiac symptoms such as chest pain or abnormal rhythms and can be performed daily following acute cardiovascular events like myocardial infarction to monitor the heart's condition.

Integrating ECG signals into the training of large language models (LLMs) can lead to the development of models that enhance diagnostic accuracy, improve workflow efficiency, support personalized patient care, and foster better communication between patients and healthcare providers. This integration allows for more nuanced interpretations of ECG data and facilitates the identification of subtle patterns that may indicate underlying health issues. \cite{quer2024potential}

The PTB-XL and MIMIC-IV-ECG datasets are key resources for ECG analysis in machine learning. The PTB-XL dataset\cite{wagner2020ptb} consists of 21,837 12-lead ECG recordings from 18,869 patients, each lasting 10 seconds. It includes annotations from up to two cardiologists, resulting in a multi-label dataset with 71 ECG statements that cover various diagnostic categories. This dataset is publicly available and designed to facilitate the training and evaluation of automatic ECG interpretation algorithms. The MIMIC-IV-ECG dataset\cite{gow2023mimic} is part of the broader MIMIC-IV critical care database, containing over 1 million ECG recordings from more than 60,000 patients. It features a wide range of clinical conditions and includes both waveform data and clinical notes, making it valuable for research in ECG signal processing and machine learning applications. Both datasets are instrumental for developing robust machine learning models in the field of cardiology.

\subsubsection{\textbf{Omics}}\

Omics refers to a comprehensive approach in biological research that examines the complete set of biological molecules within specific categories, such as genomes, transcriptomes, proteomes, and metabolomes. This approach leverages high-throughput technologies to measure these molecules simultaneously, enhancing our understanding of biological functions and interactions. In bioinformatics, various machine learning models (MLLMs) utilizing omics datasets have been proposed\cite{liu2024large} for applications like predicting genome-wide variant effects, identifying DNA-protein interactions, analyzing RNA sequences for splicing predictions, and assessing protein sequences for structural properties and functionalities. Integrating omics data into medical MLLMs enables the analysis of complex datasets to uncover novel biomarkers, predict patient responses to treatments, identify genetic markers linked to specific diseases, and facilitate personalized medicine. For example, in oncology, MLLMs can analyze genetic mutations to inform treatment strategies, while in cardiology, they help identify genetic risk factors for heart disease.
The Cancer Genome Atlas (TCGA) dataset exemplifies this approach. Launched in 2006 as a collaboration between the National Cancer Institute (NCI) and the National Human Genome Research Institute (NHGRI), TCGA includes over 20,000 tumor and normal samples from more than 11,000 patients across 33 cancer types. This dataset provides valuable insights into cancer progression, treatment responses, and genomic alterations while supporting advanced bioinformatics workflows through standardized clinical data. Another notable resource is the Human Protein Atlas (HPA) dataset, which catalogs protein and RNA expression profiles across various human tissues and conditions using high-throughput techniques like antibody-based assays and transcriptomics. The HPA encompasses data on over 17,000 unique proteins, detailing their spatial distributions, interactions, and profiles in diseases and immune cells.

Until now, we have provided a comprehensive overview of benchmark datasets across various data types, formats, and medical modalities. In the upcoming sections, we will explore how some of these datasets can be utilized for training models in specific tasks related to large language models (LLMs), categorized into discriminative and generative tasks.

\begin{table*}[htbp]
\centering
\caption{Overview of other modalities benchmark datasets in medical AI(* The abbreviation of the dataset is used as its name)}
\begin{tabular}{p{0.3\textwidth}  p{0.2\textwidth}  p{0.2\textwidth} p{0.2\textwidth}}
\hline
\textbf{Type} & \textbf{Benchmarks (Year)} \\
\hline

\textbf{Video}
&\textbf{Surgical videos:} \\
&Cholec80(2016)\cite{twinanda2016endonet}
&CholecT50(2022)\cite{nwoye2022rendezvous}
&Endoscapes (2023)\cite{murali2023endoscapes}\\
&HeiChole(2023)\cite{wagner2023comparative}
&JIGSAWS(2014)\cite{gao2014jhu}
&MedVidQA(2023)\cite{gupta2023dataset}\\
&MultiBypass140(2024)\cite{lavanchy2024challenges}
&OphNet(2024)\cite{hu2024ophnet}\\ \\

&\multicolumn{3}{l}{\textbf{Patient interaction videos:}} \\
&The Bristol Archive Project(2017)\cite{jepson2017one}\\ \\

&\multicolumn{3}{l}{\textbf{Video-based medical imaging:}} \\
&\textbf{Ultrasound:} \\
&COVID-BLUES (2024)
&POCUS (2020)\cite{born2020pocovid}
&EchoNet-Dynamic (2020)\cite{ouyang2020video}
\\ 

&\multicolumn{3}{l}{\textbf{Colonoscopy \& Endoscopy:}} \\
&ASU-Mayo (2015)\cite{tajbakhsh2015automated}
&C3VD(2023)\cite{bobrow2023}
&CVC-ClinicVideoDB(2015)\cite{bernal2015wm}\\
&EndoTect(2020)\cite{githubGitHubSimulaicprendotect2020}
&GastroLab Dataset\cite{gastrolabGastrolabImage}
&HyperKvasir(2020)\cite{borgli2020hyperkvasir}\\
&LDPolypVideo(2021)\cite{ma2021ldpolypvideo}
&REAL-Colon(2024)\cite{biffi2024real}
&S.U.N. Colonoscopy(2021)\cite{misawa2021development}\\

&\textbf{Human Posture videos:} \\
&MVOR(2018)\cite{srivastav2018mvor}
&MPII(2014)\cite{andriluka20142d}
&PoseTrack(2018)\cite{andriluka2018posetrack}\\\\

\textbf{Audio}
&\textbf{Medical classification:} \\
&HF\_Lung\_V1 (2021)\cite{hsu2021benchmarking}
&ICBHI2017 (2017)\cite{rocha2019open}
&SPRSound (2022)\cite{zhang2022sprsound}\\
&smarty4covid (2022)\cite{zarkogianni2023smarty4covid}
&PSG-Audio (2021)\cite{korompili2021psg}

\\ \\

&\textbf{Medical conversation:} \\
&Kaggle\_MSTI* (2018)\cite{kaggleMedicalSpeech}
&OGCE* (2022)\cite{fareez2022dataset}
\\ \\

\textbf{ECG}
&Apnea-ECG (2000)\cite{penzel2000apnea}
&ESI (2024)\cite{yu2024ecg} 
&ECG-QA (2024)\cite{oh2024ecg} \\
&Icentia11k (2019)\cite{tan2019icentia11k}
&MIMIC-IV-ECG (2023)\cite{gow2023mimic}
&PTB-XL (2020)\cite{wagner2020ptb} \\
&MIT-BIH(2001)\cite{moody2001impact}
&Code-15(2020)\cite{ribeiro2020automatic}
\\ \\

\textbf{Omics}
&Basset(2016)\cite{kelley2016basset}
&ENCODE(2012)\cite{encode2012integrated}
&GDC(2016)\cite{cancerNextGeneration}\\
&GTEx(2020)\cite{gtex2020gtex}
&GIAB (updated on 2024)\cite{lee2023benchmarking}
&HMDB(updated on 2022)\cite{wishart2022hmdb}\\
&HPA(2015)\cite{uhlen2015human}
&KEGG(1995)\cite{kanehisa2016kegg}
&NCBI GEO(2002)\cite{edgar2002gene}\\
&NCI-60(2023)\cite{workman2023nci}
&NSCLC-Radiomics-Genomics(2015)\cite{Aerts2015-ky}
&PDB(2019)\cite{wwpdb2019protein}\\
&TCGA(2006)\cite{tcga}
\\ \\

\hline
\end{tabular}
\label{tab:OthMod}
\end{table*}

\section{Discriminative Biomedical Tasks \& Benchmark}

Discriminative biomedical tasks and benchmarks are integral components of biomedical natural language processing (NLP), focusing on developing methods to identify and classify specific biomedical entities and their relationships. These tasks encompass various domains, such as Named Entity Recognition (NER), Relation Extraction (RE), Text Classification, Natural Language Inference, Semantic Textual Similarity, Information Retrieval, and Entity Linking. Benchmarks in this field are carefully curated datasets used to evaluate and advance the performance of machine learning models on these discriminative tasks. Each benchmark offers unique challenges that reflect real-world complexities, including identifying diseases, chemicals, proteins, adverse drug events, and gene-disease interactions, among other biomedical concepts. The development and use of these benchmarks facilitate progress in creating accurate, robust, and efficient models capable of understanding and processing biomedical text for research, clinical applications, and healthcare informatics. In this section, we will briefly discuss some commonly used datasets, as shown in Table \ref{tab:discriminative},for these types of tasks.

\begin{table*}[htbp]
\centering
\caption{Discriminative Biomedical Tasks \& Benchmark in Medical AI}
\label{tab:discriminative}
\begin{tabular}{p{0.3\textwidth} p{0.6\textwidth}}
\hline
\textbf{Task} & \textbf{Benchmarks (Year)} \\
\hline
\textbf{Named Entity Recognition (NER)} &
\begin{itemize}
    \item NCBI Disease (2014) \cite{dougan2014ncbi}
    \item JNLPBA (2004) \cite{collier2004introduction}
    \item GENIA (2003) \cite{kim2003genia}
    \item BC5CDR (2016) \cite{li2016biocreative}
    \item BC4CHEMD (2015) \cite{krallinger2015chemdner}
    \item BioRED (2022) \cite{luo2022biored}
    \item NLM-Chem-BC7 (2022) \cite{islamaj2022nlm}
    \item 2014 i2b2/UTHealth (Track 1) (2015) \cite{stubbs2015annotating}
    \item 2018 n2c2 (Track 2) (2020) \cite{henry20202018}
    \item Cadec (2015) \cite{karimi2015cadec}
    \item DDI (2013) \cite{herrero2013ddi}
    \item EU-ADR (2012) \cite{van2012eu}
    \item BioCreative VII Challenge, 2021 (2021) \cite{xu2021bch}
    \item 2010 i2b2/VA (2011) \cite{uzuner20112010}
\end{itemize} \\
\textbf{Relation Extraction} &
\begin{itemize}
    \item BC5CDR (2016) \cite{li2016biocreative}
    \item BioRED (2022) \cite{luo2022biored}
    \item ADE (2012) \cite{gurulingappa2012development}
    \item 2018 n2c2 (Track 2) (2020) \cite{henry20202018}
    \item 2010 i2b2/VA (2011) \cite{uzuner20112010}
    \item ChemProt (2017) \cite{krallinger2017overview}
    \item GDA (2019) \cite{wu2019renet}
    \item DDI (2013) \cite{segura2013semeval}
    \item GAD (2004) \cite{becker2004genetic}
    \item PGR (2019) \cite{sousa2019silver}
    \item EU-ADR (2012) \cite{van2012eu}
    \item 2014 i2b2/UTHealth (Track 2) (2015) \cite{stubbs2015annotating}
\end{itemize} \\
\textbf{Text Classification} &
\begin{itemize}
    \item ADE (2012) \cite{gurulingappa2012development}
    \item HoC (2016) \cite{baker2016automatic}
    \item OHSUMED (1994) \cite{hersh1994ohsumed}
    \item WNUT-2020 Task 2 (2020) \cite{nguyen2020wnut}
    \item Medical Abstracts (2022) \cite{schopf2022evaluating}
\end{itemize} \\
\textbf{Natural Language Inference} &
\begin{itemize}
    \item MedNLI (2018) \cite{romanov2018lessons}
    \item BioNLI (2022) \cite{bastan2022bionli}
\end{itemize} \\
\textbf{Semantic Textual Similarity} &
\begin{itemize}
    \item MedSTS (2018) \cite{wang2020medsts}
    \item 2019 n2c2/OHNLP (2020) \cite{wang20202019}
    \item BIOSSES (2017) \cite{souganciouglu2017biosses}
\end{itemize} \\
\textbf{Information Retrieval} &
\begin{itemize}
    \item TREC-COVID (2020) \cite{voorhees2021trec}
    \item NFCorpus (2016) \cite{boteva2016full}
    \item BEIR (2021) \cite{thakur2021beir}
    \item OHSUMED (1994) \cite{hersh1994ohsumed}
\end{itemize} \\
\textbf{Entity Linking/Normalization} &
\begin{itemize}
    \item NCBI Disease (2014) \cite{dougan2014ncbi}
\end{itemize} \\
\textbf{Temporal Information Extraction} &
\begin{itemize}
    \item 2012 i2b2 (2013) \cite{sun2013evaluating}
\end{itemize} \\
\textbf{Assertion Classification} &
\begin{itemize}
    \item 2010 i2b2/VA (2011) \cite{uzuner20112010}
\end{itemize} \\
\textbf{Multi-Task} &
\begin{itemize}
    \item CMeEE (2021) \cite{zhang2021cblue}
    \item MIMIC-III (2016) \cite{johnson2016mimic}
\end{itemize} \\
\hline
\end{tabular}
\label{tab:dis}
\end{table*}

\textit{Named Entity Recognition (NER)} focuses on identifying and classifying key information within text, such as diseases, chemicals, genes, and drugs, making it a cornerstone task in biomedical text mining. In the medical domain, NER plays a vital role in structuring vast amounts of unstructured biomedical data, enabling more efficient information retrieval, patient record analysis, and biomedical research. One of the significant applications of NER in biomedicine is disease name recognition and concept normalization, as exemplified by the NCBI Disease corpus\cite{dougan2014ncbi}. Datasets like GENIA\cite{kim2003genia} and JNLPBA\cite{collier2004introduction} extend NER's capabilities for bio-text mining and bio-entity recognition. The BC5CDR\cite{li2016biocreative} and BC4CHEMD\cite{krallinger2015chemdner} corpora focus on the extraction of chemical-disease relations and chemical information, respectively, while NLM-Chem-BC7\cite{islamaj2022nlm} provides full-text biomedical annotations for chemical entities. The 2014 i2b2/UTHealth (Track 1)\cite{stubbs2015annotating} dataset addresses the de-identification of clinical narratives, ensuring patient privacy. Additionally, the Cadec corpus\cite{karimi2015cadec} highlights adverse drug event extraction from social media, and DDI\cite{segura2013semeval} targets drug-drug interaction identification. These corpora serve as benchmarks for advancing NER systems in biomedical informatics.

\textit{Relation Extraction} tasks are designed to identify and extract semantic relationships between entities within biomedical texts. In the context of biomedical research, relation extraction plays a crucial role in linking different biological entities—such as genes, proteins, diseases, and chemicals—to uncover meaningful connections, thereby facilitating the understanding of complex biological processes and aiding in clinical decision-making. Several datasets have been developed to support and advance this task. BC5CDR\cite{li2016biocreative} targets chemical-disease relations with high-quality annotations, while BioRED\cite{luo2022biored} provides a diverse set of biomedical entities and relations to improve model robustness. The ADE\cite{gurulingappa2012development} and 2018 n2c2 (Track 2) datasets\cite{henry20202018} focus on extracting drug-related adverse effects and medication information from clinical narratives, highlighting challenges in ADE\cite{gurulingappa2012development} identification. 2010 i2b2/VA\cite{uzuner20112010} covers the broader extraction of medical concepts, assertions, and relations in clinical text. ChemProt\cite{krallinger2017overview} addresses chemical-protein interactions, vital for pathway analysis and drug discovery. Gene-disease associations are captured in the GDA dataset\cite{wu2019renet}, enhancing extraction from biomedical literature, and DDI\cite{herrero2013ddi} focuses on identifying drug-drug interactions, crucial for pharmacological safety. Lastly, GAD\cite{becker2004genetic} and PGR datasets\cite{sousa2019silver} explore genetic associations and phenotype-gene relations, contributing to genomics research. Together, these datasets support the development of models that capture and understand complex biomedical relations.

\textit{Text classification} in the biomedical domain is pivotal for organizing, categorizing, and extracting meaningful insights from vast biomedical text sources. These tasks encompass a range of applications, from identifying informative content on social media during health crises to categorizing scientific literature for specific research themes. Notable datasets for text classification include WNUT-2020 Task 2\cite{nguyen2020wnut}, which focused on identifying informative COVID-19 tweets, and OHSUMED\cite{hersh1994ohsumed}, designed for developing information retrieval systems in medical literature. The Medical Abstracts dataset\cite{schopf2022evaluating} facilitated advancements in unsupervised text classification, while ADE\cite{gurulingappa2012development} served as a benchmark for extracting drug-related adverse effects from case reports. The HoC dataset\cite{baker2016automatic}, on the other hand, categorizes literature based on cancer hallmarks, enhancing targeted research. These datasets collectively support advancements in biomedical research, improving knowledge dissemination and aiding clinical decision-making.

\textit{Natural Language Inference (NLI)} is a task focusing on determining the inferential relationship between two text segments, typically referred to as a "premise" and a "hypothesis." In the biomedical field, NLI plays a critical role in interpreting and validating clinical and scientific statements, enhancing the understanding of medical narratives and supporting evidence-based practice. Two key datasets that contribute to this task in biomedical contexts are MedNLI\cite{romanov2018lessons} and BioNLI\cite{bastan2022bionli}. MedNLI\cite{romanov2018lessons} is a dataset designed for NLI in the clinical domain, annotated by clinicians and grounded in medical history, emphasizing the use of transfer learning and domain-specific knowledge to enhance model performance. BioNLI\cite{bastan2022bionli}, meanwhile, introduces a biomedical NLI dataset generated with lexico-semantic constraints to create adversarial examples, providing a diverse set of challenges for benchmarking NLI models. Both datasets underscore the need for specialized resources to address the complexities of medical language, focusing on robust model development through domain adaptation and challenging negative example creation to improve inference in biomedical texts.

\textit{Semantic Textual Similarity (STS)} in the biomedical field focuses on quantifying the degree of similarity between pairs of clinical or biomedical sentences. The task plays a crucial role in supporting applications like clinical decision support, question answering, and information retrieval within the medical domain. Notable datasets such as 2019 n2c2/OHNLP\cite{wang20202019}, BIOSSES\cite{souganciouglu2017biosses}, and MedSTS\cite{wang2020medsts} have been developed to benchmark and stimulate research in this area. The 2019 n2c2/OHNLP\cite{wang20202019} shared task demonstrated the effectiveness of neural language models for clinical text understanding, while BIOSSES\cite{souganciouglu2017biosses} addressed the limitations of general-domain similarity systems in handling biomedical text by integrating string similarity, ontology-based, and supervised methods. MedSTS\cite{wang2020medsts}, with its expert-annotated corpus of clinical sentence pairs, enables the evaluation of STS systems in medical contexts. Together, these datasets drive progress in refining semantic similarity models tailored to biomedical language.

\textit{Information Retrieval (IR)} tasks in the biomedical domain are crucial for efficiently accessing and organizing vast amounts of biomedical literature and data, particularly to support clinical decision-making, research, and public health initiatives. IR tasks often aim to develop systems capable of understanding and retrieving relevant information from biomedical texts, catering to the evolving needs of researchers and clinicians. Several notable datasets have advanced IR in biomedicine. TREC-COVID\cite{voorhees2021trec} was developed to address the evolving information needs during the COVID-19 pandemic, aiming to build a robust infrastructure for pandemic-related search. The NFCorpus\cite{boteva2016full} focuses on learning-to-rank within medical texts, emphasizing patient-centered queries for improved retrieval performance. BioASQ\cite{krithara2023bioasq}, part of the BEIR\cite{thakur2021beir} benchmark, evaluates IR model generalizability across diverse biomedical queries, revealing that re-ranking and late-interaction models can outperform baselines like BM25, albeit at a higher computational cost. Lastly, OHSUMED\cite{hersh1994ohsumed} serves as a foundational resource for IR system evaluation, providing a comprehensive test collection that has guided the development and assessment of IR systems in the medical field.

\textit{Entity Linking/Normalization} is a critical task within biomedical text processing that involves mapping textual mentions of biomedical entities to standardized concepts within a controlled vocabulary. This task enhances the interpretability and interoperability of biomedical data by linking varied expressions to unified concepts, aiding both research and clinical applications. In the biomedical domain, linking entities like diseases, drugs, or genes to standardized ontologies facilitates data integration, retrieval, and downstream analyses, such as relation extraction or question answering. Several datasets are pivotal for Entity Linking/Normalization tasks, offering benchmarks and resources for evaluating model performance. For instance, the NCBI Disease dataset\cite{dougan2014ncbi} includes disease mentions mapped to Medical Subject Headings (MeSH) and Online Mendelian Inheritance in Man (OMIM) concepts, aiding disease name recognition and concept normalization. It consists of 6,892 disease mentions linked to 790 unique disease concepts, with most mapped to MeSH. The dataset’s benchmarking experiments show an F-measure of 63.7\% for the best-performing normalization method, highlighting the challenges in linking textual mentions to structured disease concepts. Such datasets enhance the precise identification and normalization of biomedical entities, supporting cohesive research and applications in disease diagnosis, drug discovery, and personalized medicine.

\textit{Temporal Information Extraction} is a crucial task within the broader scope of extracting and understanding clinical and biomedical narratives. It focuses on identifying and analyzing temporal information present in clinical texts, such as events, temporal expressions (e.g., dates or durations), and temporal relations that delineate the chronological sequence and co-occurrence of medical events. This enables more accurate tracking of disease progression, treatment effects, and patient outcomes over time. An example within the biomedical domain is the 2012 i2b2 dataset\cite{uzuner20112010}, which evaluates systems for temporal relation extraction in clinical narratives. It covers tasks like event detection, temporal expression extraction, normalization, and classification of temporal relationships between events. The results indicate that combining statistical machine learning and rule-based approaches enhances the performance of temporal information extraction. This task is vital for improving temporal reasoning in clinical texts, facilitating better decision-making, and enhancing patient care, although challenges remain in refining annotation guidelines and techniques for complex expressions.

\textit{Assertion classification} within biomedical contexts focuses on determining the factuality or status of clinical concepts mentioned in texts. This task is essential for understanding how medical information is conveyed in clinical narratives, distinguishing whether a concept (such as a disease, symptom, or treatment) is affirmed, negated, hypothetical, or uncertain. Its primary objective is to enrich the context of named entities or relationships by categorizing them based on their associated assertions. In practical applications, assertion classification has been crucial in refining clinical information extraction from electronic health records (EHRs). For instance, the “2010 i2b2/VA” dataset\cite{uzuner20112010} was used in the i2b2 challenge, where assertion classification aimed to interpret clinical concepts by identifying their nature (e.g., presence, absence, or uncertainty). This task is fundamental for understanding clinical notes and supporting patient data analysis. Although assertion classification has been extensively studied, the variability and complexity of clinical language necessitate continuous advancement to improve accuracy and contextual interpretation.

In addition, benchmarks designed for \textit{Multi-task} learning have gained increasing attention as they provide a holistic framework to assess and develop models capable of handling diverse biomedical tasks concurrently. These benchmarks aim to challenge machine learning models to learn and generalize across multiple related tasks, reflecting the complexities inherent in real-world medical applications where multiple forms of data and objectives often coexist. For instance, the CBLUE benchmark\cite{zhang2021cblue} offers a range of tasks for Chinese biomedical language understanding, challenging multi-task models with tasks like information extraction and classification. Similarly, MIMIC-III\cite{johnson2016mimic}, a comprehensive critical care database, supports a spectrum of clinical NLP applications from temporal event detection to patient state classification, promoting the development of integrative models. Such benchmarks exemplify the value of multi-task learning in biomedical NLP, driving progress toward models that can handle the multifaceted nature of clinical narratives and research data.
 
\section{Generative Benchmarks}

In recent years, the advancement of Medical Large Language Models (MLLMs) has been significantly propelled by the development of diverse and specialized benchmarks that evaluate a wide array of tasks pertinent to the medical domain. These benchmarks not only provide a means to assess the performance of MLLMs but also guide research efforts towards addressing critical challenges in medical artificial intelligence. The tasks encompass areas such as medical report generation, synthetic data creation, clinical summarization, question answering, and more—each playing a vital role in enhancing healthcare delivery and medical research. In this section, we will briefly discuss some commonly used datasets, as shown in Table \ref{tab:gen},for these types of tasks.

\begin{table*}[htbp]
\centering
\caption{Generative Benchmarks in Medical AI}
\begin{tabular}{p{0.3\textwidth} p{0.6\textwidth}}
\hline
\textbf{Task} & \textbf{Benchmarks (Year)} \\
\hline
\textbf{Medical Report Generation} &
\begin{itemize}
    \item MIMIC-CXR (2019)\cite{johnson2019mimic}
    \item IU X-Ray (2015)\cite{demner2016preparing}
    \item PEIR Digital Library (2013)\cite{pavlopoulos2019survey}
    \item OpenI Chest X-rays (2012)\cite{nihOpeniNgWeb}
    \item CheXpert (2019)\cite{irvin2019chexpert}
\end{itemize} \\
\textbf{Synthetic Data Generation} &
\begin{itemize}
    \item MIMIC-III (2016)\cite{johnson2016mimic}, MIMIC-IV (2020)\cite{johnson2023mimic}
    \item n2c2 Clinical NLP Challenges (2006--present)\cite{stubbs2019cohort}
    \item CT-GAN (2018)\cite{mirsky2019ct}
    \item BraTS (2012)\cite{baid2021rsna}
    \item NIH Chest X-ray Dataset (2017)\cite{wang2017chestx}
\end{itemize} \\
\textbf{Medical Image Synthesis} &
\begin{itemize}
    \item LIDC-IDRI (2011)\cite{https://doi.org/10.7937/k9/tcia.2015.lo9ql9sx}
    \item BraTS (2012)\cite{baid2021rsna}
    \item NIH Chest X-rays (2017)\cite{wang2017chestx}
    \item CheXpert (2019)\cite{irvin2019chexpert}
\end{itemize} \\
\textbf{Predictive Text Generation for Clinical Decision Support} &
\begin{itemize}
    \item MIMIC-III (2016)\cite{johnson2016mimic}, MIMIC-IV (2020)\cite{johnson2023mimic}
    \item eICU Collaborative Research Database (2018)\cite{pollard2018eicu}
    \item PhysioNet Challenge (2000--present)\cite{goldberger2000physiobank}
    \item TREC Clinical Decision Support Track (2014)\cite{hovy2001toward}
\end{itemize} \\
\textbf{Clinical Summarization} &
\begin{itemize}
    \item n2c2 Clinical NLP Summarization Task (2014, 2018)\cite{stubbs2019cohort}
    \item MEDIQA Challenge (2019)\cite{abacha2019overview}
    \item PubMed Dataset (ongoing since 1996)\cite{roberts2001pubmed}
    \item CORD-19 (2020)\cite{wang-etal-2020-cord}
    \item PMC (PubMed Central) (2000)\cite{roberts2001pubmed}
    \item MentSum (2021)\cite{sotudeh-etal-2022-mentsum}
    \item MeQSum (2016)\cite{abacha2019summarization}
    \item MedQSum (2020)\cite{10373720}
\end{itemize} \\
\textbf{Dialogue Generation for Patient Interaction} &
\begin{itemize}
    \item MEDIQA-QA (2019)\cite{abacha2019overview}
    \item EmpatheticDialogues Dataset (2019)\cite{rashkin2018towards}
    \item Medical Virtual Assistant Dialogue Corpora (various years)\cite{li2022}
\end{itemize} \\
\textbf{Educational Content Generation} &
\begin{itemize}
    \item MSQA (Medical Student Question Answering) (2021)
    \item USMLE datasets (derived from exams dating back to 1992)\cite{usmleHomeUnited}
    \item MedQA (2020)\cite{jin2021disease}
\end{itemize} \\
\textbf{Text Simplification} &
\begin{itemize}
    \item MultiCochrane (2022)\cite{joseph2023multilingual}
    \item AutoMeTS (2019)\cite{van-etal-2020-automets}
\end{itemize} \\
\textbf{Text Summarization} &
\begin{itemize}
    \item PubMed (since 1996)\cite{roberts2001pubmed}
    \item PMC (2000)\cite{roberts2001pubmed}
    \item CORD-19 (2020)\cite{wang-etal-2020-cord}
    \item MentSum (2021)\cite{sotudeh-etal-2022-mentsum}
    \item MeQSum (2016)\cite{abacha2019summarization}
    \item MedQSum (2020)\cite{10373720}
\end{itemize} \\
\hline
\end{tabular}
\label{tab:gen}
\end{table*}

One of the foundational tasks in this sphere is Medical Report Generation, which focuses on the automatic creation of detailed medical reports from raw clinical data, notably medical images. This task is essential for improving clinical workflows and reducing the burden on healthcare professionals by providing consistent and comprehensive reports. Datasets like MIMIC-CXR\cite{johnson2019mimic} and IU X-Ray\cite{demner2016preparing} are pivotal in this regard. MIMIC-CXR\cite{johnson2019mimic} offers a vast collection of over 370,000 chest X-ray images accompanied by radiology reports, serving as a critical resource for training models to generate diagnostic narratives from imaging data. Similarly, the IU X-Ray dataset\cite{demner2016preparing} comprises approximately 8,000 chest X-ray images paired with radiology reports from the Indiana University hospital network, providing a benchmark for evaluating report generation models on a manageable scale. Additional resources like the PEIR Digital Library\cite{pavlopoulos2019survey} and OpenI Chest X-rays\cite{nihOpeniNgWeb} contribute diverse imaging case studies and associated reports, enriching the training data necessary for developing robust medical report generation systems. The CheXpert dataset\cite{irvin2019chexpert}, while primarily used for classification tasks, also includes radiology reports that can be utilized to enhance report generation capabilities.

Parallel to report generation is the task of Synthetic Data Generation, which addresses issues related to data scarcity and privacy by creating artificial datasets that simulate real clinical data. The utilization of synthetic data is crucial for training machine learning models without compromising patient confidentiality. Datasets such as MIMIC-III\cite{johnson2016mimic} and MIMIC-IV\cite{johnson2023mimic}, which contain extensive de-identified health records, are often used to generate synthetic clinical text and time-series data. The n2c2 Clinical NLP Challenges\cite{stubbs2019cohort} also incorporate synthetic data to test the adaptability of natural language processing (NLP) models to artificial yet clinically relevant text. In the realm of image synthesis, CT-GAN methodologies\cite{mirsky2019ct} facilitate the generation of synthetic medical images, with datasets like BraTS (Brain Tumor Segmentation)\cite{baid2021rsna} and the NIH Chest X-ray Dataset\cite{wang2017chestx} serving as benchmarks for synthetic image generation, particularly in producing brain tumor images and chest X-rays, respectively.

Expanding upon synthetic data, Medical Image Synthesis is a task dedicated to generating realistic medical images using computational models, which is crucial for data augmentation and enhancing diagnostic models. The LIDC-IDRI dataset\cite{https://doi.org/10.7937/k9/tcia.2015.lo9ql9sx} provides lung CT scans with annotated lesions, serving as a benchmark for lung cancer image generation. Similarly, the BraTS dataset\cite{baid2021rsna} continues to be instrumental in synthesizing brain MRI images for tumor segmentation tasks. The NIH Chest X-rays\cite{wang2017chestx} and CheXpert datasets\cite{irvin2019chexpert} are utilized for image synthesis and enhancement, enabling the creation of synthetic chest X-ray images to augment training datasets and improve model robustness against class imbalances.

In the context of clinical decision-making, Predictive Text Generation for Clinical Decision Support is a task that involves creating models capable of predicting clinical outcomes, recommending treatments, or providing diagnostic suggestions based on patient data. The rich data from MIMIC-III\cite{johnson2016mimic}, MIMIC-IV\cite{johnson2023mimic}, and the eICU Collaborative Research Database\cite{pollard2018eicu} are pivotal for training such predictive models. These datasets enable the development of systems that can, for instance, anticipate patient deterioration or suggest medication adjustments, thereby assisting healthcare professionals in making informed decisions. The PhysioNet Challenge\cite{goldberger2000physiobank} provides various datasets for clinical event prediction tasks, such as mortality prediction and arrhythmia classification, while the TREC Clinical Decision Support Track\cite{hovy2001toward} offers benchmarks for retrieving clinical evidence to support decision-making.

Another significant task is Clinical Summarization, which aims to condense lengthy and complex medical documents into concise summaries that highlight the most critical information. This task is essential for improving information accessibility and aiding quick decision-making. Datasets like the n2c2 Clinical NLP Summarization\cite{stubbs2019cohort} Task provide resources for summarizing patient records, focusing on extracting relevant information for clinical use. The MEDIQA Challenge\cite{abacha2019overview} offers datasets for clinical text summarization and question-answering, promoting the development of systems capable of generating coherent and relevant summaries of medical texts. Extensive resources like the PubMed Dataset and PMC (PubMed Central)\cite{roberts2001pubmed} provide vast collections of biomedical research articles for summarization, aiding researchers and clinicians in staying updated with the latest findings. During the COVID-19 pandemic, the CORD-19 dataset\cite{wang-etal-2020-cord} became crucial for summarizing rapidly evolving research on the virus. Additionally, datasets like MentSum\cite{sotudeh-etal-2022-mentsum}, MeQSum\cite{abacha2019summarization}, and MedQSum\cite{10373720} focus on summarizing mental health posts and consumer health questions, emphasizing the importance of patient-centered communication.

The development of conversational agents in healthcare is encapsulated in the task of Dialogue Generation for Patient Interaction. This involves creating systems capable of engaging in natural and informative conversations with patients, which is significant for virtual assistants, telemedicine applications, and tools that support patient education. The MEDIQA-QA\cite{abacha2019overview} dataset provides a foundation for medical question-answering and patient dialogues, focusing on generating accurate responses to patient inquiries. The EmpatheticDialogues Dataset\cite{rashkin2018towards}, though not exclusively medical, is adapted to train models that can interact empathetically with patients, enhancing the quality of patient-provider communication. Moreover, specialized Medical Virtual Assistant Dialogue Corpora \cite{li2022}are used to train models in understanding and effectively responding to patient queries.

Supporting medical education, the task of Educational Content Generation involves developing models that can produce informative and accurate educational materials in medicine. Datasets such as MSQA (Medical Student Question Answering Dataset), derived from medical student assessments, evaluate models on medical question answering, thereby aiding in the development of educational tools for students. The USMLE (United States Medical Licensing Examination) dataset\cite{usmleHomeUnited}, comprising examination questions, is utilized to train models that generate or answer questions at the level required for medical licensing, supporting self-assessment and study. Similarly, MedQA\cite{jin2021disease} provides a large-scale question-answering dataset from medical exams, essential for developing models that can assist in medical education by generating practice questions and explanations.

Addressing the complexity of medical language, the task of Abbreviation Expansion and Clinical Normalization focuses on interpreting and standardizing medical abbreviations and terms within clinical texts. This is crucial for ensuring clear communication and reducing ambiguity in medical records. The n2c2 Challenges\cite{jin2021disease} include tasks on abbreviation disambiguation and clinical normalization, offering datasets to train models that interpret shorthand in clinical narratives accurately. The Medical Abbreviation Disambiguation Dataset further supports this task by providing instances of medical abbreviations with their expanded forms.

Multimodal Translation Tasks in medicine involve integrating multiple types of data, such as text and images, to perform tasks like image captioning and visual question answering. This task enhances the ability of AI systems to understand and interpret complex medical information presented in different formats. The ImageCLEFmed benchmark dataset\cite{de2016overview} offers challenges for multimodal medical information retrieval, requiring models to interpret both images and associated text. VQA-Med (Visual Question Answering in Medicine)\cite{ben2019vqa,ben2021overview} focuses on answering questions based on medical images, combining image analysis with natural language understanding. The MedNLI dataset\cite{romanov2018lessons}, designed for textual inference tasks in medical reasoning, aids models in understanding and reasoning with clinical text.

A core component of medical AI is the task of Question Answering, which involves developing systems that can provide accurate and relevant answers to medical questions posed by clinicians, researchers, or patients. This task is fundamental in supporting information retrieval and clinical decision-making. The Huggingface Leaderboard evaluates medical language models across multiple QA tasks, providing benchmarks for performance comparison. The Open Medical-LLM Leaderboard encompasses datasets like MedQA\cite{jin2021disease}, PubMedQA\cite{jin2019pubmedqa}, MedMCQA\cite{pal2022medmcqa}, and subsets of the MMLU\cite{hendryckstest2021} related to medicine, covering a broad spectrum of medical knowledge. MedQA\cite{jin2021disease} and MedMCQA \cite{pal2022medmcqa}are derived from medical examinations and provide large-scale question-answering datasets essential for training models on complex medical queries. PubMedQA\cite{jin2019pubmedqa} focuses on biomedical research questions, while BioASQ-QA\cite{krithara2023bioasq} offers a biomedical QA benchmark with factoid and summary answers. Datasets like emrQA\cite{pampari2018emrqa} and CliCR\cite{vsuster2018clicr} address QA in electronic medical records and clinical case report comprehension, respectively. Specialized datasets like COVID-QA and Health-QA provide resources for QA related to COVID-19 and general healthcare queries, highlighting the adaptability of models to current and diverse information needs.

In the interest of making complex medical information more accessible, Text Simplification is a task that involves translating intricate medical texts into language that is easier for non-experts to understand. This is critical for improving health literacy and patient engagement. Datasets such as MultiCochrane\cite{joseph2023multilingual} provide multilingual resources for simplifying medical texts from Cochrane reviews, enabling models to produce simpler versions of complex medical literature. The AutoMeTS\cite{van-etal-2020-automets} dataset supports the development of autocomplete systems for medical text simplification, assisting in writing simplified medical content.

Lastly, the task of Text Summarization in the medical domain involves condensing lengthy medical documents into brief summaries that capture essential information. This is vital for aiding healthcare professionals in quickly assimilating information and supports patients in understanding medical content. Extensive datasets like PubMed and PMC \cite{roberts2001pubmed} serve as rich resources for developing summarization models. The CORD-19 dataset has been particularly important during the COVID-19 pandemic for summarizing research articles related to the virus. Datasets focusing on mental health and patient inquiries, such as MentSum\cite{sotudeh-etal-2022-mentsum}, MeQSum\cite{abacha2019summarization}, and MedQSum\cite{10373720}, contribute to the development of tools that summarize mental health posts and consumer health questions, emphasizing personalized communication.

In conclusion, the array of tasks and associated benchmarks in medical AI reflects the multifaceted challenges and opportunities within the field. These datasets are indispensable for the continued progress of MLLMs, providing the foundation upon which models are trained, evaluated, and refined. By addressing tasks ranging from report generation and data synthesis to question answering and text simplification, these benchmarks collectively drive innovation aimed at enhancing healthcare delivery, supporting clinical decision-making, and improving patient engagement. The integration of these datasets into research efforts ensures that AI systems are equipped to meet the complex demands of medical practice and research, ultimately contributing to better health outcomes and advancing the field of medical artificial intelligence. 

\section{Discussion and Future Perspectives}

As numerous Multimodal Large Language Models (MLLMs) demonstrate exceptional capabilities, deep learning is advancing medical research to new heights. In the development of medical intelligence, multimodal datasets play a crucial role in enhancing MLLMs' knowledge across medical domains. We have identified a collection of multimodal datasets that encompass text, images, and omics data. However, current works face several significant limitations:\

1. Limited Language Diversity: Most text datasets used for training MLLMs are primarily in English or Chinese. This limitation hinders MLLMs' understanding of complex medical terminology in other languages and introduces implicit cultural biases.\

2. Insufficient Large-Scale Annotated Medical Image Datasets: The lack of extensive datasets in medical imaging poses a considerable challenge. Developing high-quality, domain-specific image-caption pairs requires substantial human expertise, making the creation of automatic captioning systems a persistent issue.\

3. Inadequate Variety of Data Types: Existing collections of medical datasets primarily cover foundational knowledge. However, they often fail to capture advanced concepts and the complexities inherent in real-life practices and the intricate 3D structures of biological systems. There is a pressing need to gather additional modalities, such as audio and video.\

4. Poor Structured Representation of Omics Data: Omics data, which provide rich insights into biological structures, are often underrepresented in traditional deep learning domains like computer vision and natural language processing. The absence of effective embedding methods complicates the integration of this data into MLLM training.\

5. Hallucinations: Large Language Models (LLMs) often generate outputs that are plausible but factually incorrect, a phenomenon known as hallucination. This issue arises from various factors, including data deficiencies, statistical blind spots, and limited contextual understanding\cite{peng2024securing}.\

To address these challenges and meet the growing demand for multimodal datasets that evaluate reasoning and comprehension skills in LLMs, it is essential to develop more diverse and robust benchmarks. One potential approach to curating such datasets is through augmentation using synthesis methods\cite{van2024synthetic}. Future efforts could explore synthetic data to bridge modality gaps; for instance, creating textual datasets in languages beyond English and Chinese via GPT-based translations, or generating audio data from text using text-to-speech models derived from sources like electronic health records and doctor-patient dialogues. Synthetic datasets including high-quality, fact-checked information that addresses known biases or inaccuracies in existing datasets can also be curated to migitate the hallucination problem.

\section{Conclusion}
In conclusion, the field of medical AI benchmarks is characterized by diverse and multifaceted datasets that address a wide array of tasks, each critical to advancing the capabilities of Multimodal Large Language Models (MLLMs). These benchmarks support crucial developments in medical AI by providing standardized datasets for training, evaluating, and refining models. They encompass tasks like report generation, data synthesis, question answering, and text simplification, contributing significantly to innovations aimed at improving healthcare delivery, supporting clinical decision-making, and enhancing patient outcomes.

The integration of various data modalities, from text to images and omics data, ensures that AI systems are capable of meeting the complex demands inherent in medical practice and research. This convergence of data types enables models to develop robust, context-aware reasoning capabilities essential for real-world applications. However, challenges persist, including the need for more diverse language support, better representation of complex data types, and structured approaches to omics data. Addressing these gaps through future research and benchmark development will be pivotal in harnessing the full potential of AI to transform healthcare.

Ultimately, these benchmarks play an indispensable role in the advancement of medical AI, providing the necessary resources to guide model development and ensure that AI technologies can effectively enhance medical research, practice, and patient care.

\newpage
\end{document}